\documentclass[12pt, letterpaper]{article}

\usepackage[utf8]{inputenc}
\usepackage[T1]{fontenc}
\usepackage{mathptmx} 
\usepackage{microtype}

\usepackage[
  top=1in, bottom=1in,
  left=1in, right=1in
]{geometry}

\usepackage{setspace}
\usepackage{parskip}
\usepackage{amsmath, amssymb, amsthm}

\usepackage{graphicx}
\usepackage{float}
\usepackage[labelfont=bf, font=small]{caption}
\usepackage{subcaption}

\usepackage{booktabs}
\usepackage{tabularx}
\usepackage{array}
\usepackage{multirow}
\usepackage{adjustbox}
\usepackage{makecell}

\usepackage{listings}
\usepackage{xcolor}

\definecolor{codebg}{RGB}{245,245,245}
\definecolor{codegreen}{RGB}{0,128,0}
\definecolor{codegray}{RGB}{128,128,128}
\definecolor{codepurple}{RGB}{148,0,211}
\definecolor{codeblue}{RGB}{0,0,205}

\lstdefinestyle{pythonstyle}{
backgroundcolor=\color{codebg},
commentstyle=\color{codegreen}\itshape,
keywordstyle=\color{codeblue}\bfseries,
stringstyle=\color{codepurple},
numberstyle=\tiny\color{codegray},
basicstyle=\ttfamily\small,
breakatwhitespace=false,
breaklines=true,
captionpos=b,
keepspaces=true,
numbers=left,
numbersep=8pt,
showspaces=false,
showstringspaces=false,
showtabs=false,
tabsize=4,
frame=single,
rulecolor=\color{codegray},
language=Python
}
\usepackage[
colorlinks=true,
linkcolor=black,
citecolor=blue!70!black,
urlcolor=blue!70!black,
pdftitle={Your Paper Title},
pdfauthor={Author One, Author Two}
]{hyperref}
\usepackage{url}

\usepackage[
backend=bibtex,
style=numeric,
sorting=none,
maxbibnames=6
]{biblatex}
\usepackage{lipsum} 
\usepackage{enumitem}
\usepackage{authblk}
\usepackage{lineno} 

\usepackage{titlesec}
\titleformat{\section}{\large\bfseries}{\thesection.}{0.5em}{}
\titleformat{\subsection}{\normalsize\bfseries}{\thesubsection.}{0.5em}{}
\titleformat{\subsubsection}{\normalsize\itshape}{\thesubsubsection.}{0.5em}{}

\newcommand{\toolname}{\texttt{CLABTOOLKIT}} 

\title{
\vspace{-1cm}
{\LARGE\bfseries \toolname: An Open-Source Toolkit for Routine Processing, Manipulation, and Visualization of Neuroimaging Data}\\[0.4cm]
}

\author[1,*]{Yasser Alemán-Gómez}
\author[1]{Nino Hervé}
\author[1,2]{Patric Hagmann}

\affil[1]{Connectomics Lab, Department of Radiology, Lausanne University Hospital (CHUV), Lausanne, Switzerland}
\affil[2]{University of Lausanne (UNIL), Lausanne, Switzerland}
\affil[*]{Corresponding author}

\date{} 

\begin{document}

\maketitle
\vspace{0.5cm}

\noindent
\begin{tabular}{@{}ll}
\textbf{Word count:} & \textasciitilde{}5903 words (excluding abstract, references, and captions) \\
\textbf{Figures:}& 4 \\
\textbf{Tables:} & 1 \\
\end{tabular}

\vspace{0.8cm}


\vspace{0.8cm}

\noindent\textbf{Running title:} \toolname{} — [A toolkit for routine neuroimaging processing]

\vspace{0.8cm}

\newpage

\phantomsection
\addcontentsline{toc}{section}{Abstract}

\begin{center}
{\large\bfseries Abstract}
\end{center}

Neuroimaging research requires manipulating heterogeneous data structures—not only raw magnetic resonance imaging (MRI) volumes but also volumetric parcellations, cortical surface meshes, tractograms, and connectivity matrices—across tools with incompatible interfaces and file formats, forcing researchers to repeatedly re-implement routine but technically demanding operations. We present \textsc{CLABTOOLKIT}, an open-source Python package that consolidates these operations into a single, coherent framework by representing volumetric, surface, and streamline data as interoperable Python objects. Five core data structures—\texttt{Parcellation}, \texttt{Surface}, \texttt{AnnotParcellation}, \texttt{Tractogram}, and \texttt{Connectome}—encapsulate the most common neuroanatomical entities and provide consistent methods for loading, processing, and exporting data across standard neuroimaging formats (e.g., NIfTI, GIFTI, FreeSurfer annotations, TCK/TRK), including functional and structural connectome generation directly from a parcellation and scalar-map projection onto tractogram streamlines. Complementary modules support Brain Imaging Data Structure (BIDS) dataset management, FreeSurfer integration, diffusion MRI processing, morphometric analysis, graph-theoretical network analysis, and graphics processing unit (GPU)-accelerated multi-panel visualization via PyVista. In total, the toolkit comprises 19 modules organised into six architectural layers, exposing 13 object-oriented classes with 234 methods and 207 standalone functions, and a JSON-based configuration system enables workflow customization without code modification. Unlike existing neuroimaging libraries, which typically address parcellation handling, surface visualization, or tractography separately, \textsc{CLABTOOLKIT} combines color and lookup-table management, parcellation manipulation, multi-surface visualization, and tractography utilities within a single framework. \textsc{CLABTOOLKIT} is compatible with Python 3.9–3.12 and is released under the Apache 2.0 license. Source code, documentation, and example workflows are available at \url{https://github.com/connectomicslab/clabtoolkit}.

\vspace{0.5cm}
\newpage


\section*{Introduction}
\label{sec:introduction}

The neuroimaging community has developed a rich and mature ecosystem of software tools that support a wide range of data analysis tasks. Libraries such as NiBabel \cite{brett_nipynibabel_2023} provide robust handling of neuroimaging file formats, while higher-level frameworks like Nilearn \cite{contributors_nilearn_nodate} facilitate statistical analysis and functional connectivity modeling. In parallel, visualization tools such as FURY \cite{garyfallidis_fury_2021} and PySurfer \cite{waskom_nipypysurfer_2020} enable interactive exploration of volumetric and surface-based data. Together, these tools have significantly accelerated methodological development and lowered barriers to entering complex neuroimaging analyses.\par
    
Despite this richness, the practical use of these resources remains fragmented. Neuroimaging workflows often require researchers to combine multiple libraries, each addressing a specific aspect of the analysis pipeline. For instance, one tool may be used for atlas or parcellation handling, another for surface visualization \cite{waskom_nipypysurfer_2020}, and yet another for connectivity or statistical analysis \cite{rubinov_complex_2010}. While this modularity offers flexibility, it also introduces substantial overhead: users must manage differences in data structures, coordinate systems, naming conventions, and input/output formats across tools. First, many commonly required operations—such as parsing index ranges, filtering hierarchical file structures, harmonizing region labels across atlases, handling inconsistently formatted metadata tables, or assembling multi-view visualizations—are repeatedly re-implemented across projects. Although conceptually straightforward, these tasks are often technically intricate due to inconsistencies in data representations and conventions. Second, such ad hoc implementations are rarely standardized or thoroughly tested, increasing the likelihood of subtle errors that can propagate through analyses and undermine reproducibility. Third, the need to master multiple independent tools, each with its own interface and assumptions, creates a steep learning curve for new users and limits accessibility, particularly for researchers without strong programming backgrounds. As a consequence, a considerable portion of research effort is devoted not to scientific inquiry but to implementing and maintaining the code to bridge these components.\par
    
Several recent efforts have attempted to mitigate these issues by promoting standardization (e.g., Nipype \cite{gorgolewski_nipype_2016}, the Brain Imaging Data Structure (BIDS) \cite{gorgolewski_brain_2016-1}, or fMRIPrep \cite{esteban_fmriprep_2019}) and by providing higher-level abstractions for specific domains of analysis. However, these approaches typically focus on pipelining, data organization, or specific methodological niches and do not fully address the need for a unified layer of reusable, basic tasks found in every neuroimaging processing routine. In particular, there remains a gap between general-purpose neuroimaging libraries and the practical utilities required to efficiently manipulate parcellations, coordinate surface and volumetric representations, and generate publication-ready visualizations.\par
    
To address this gap, we introduce \textsc{CLABTOOLKIT}, an open-source Python package that consolidates essential neuroimaging utilities into a single, coherent framework. The toolkit provides a consistent interface for common operations, including parcellation manipulation, region selection and indexing, surface-based processing, and flexible visualization workflows. By abstracting away repetitive and error-prone implementation details, \textsc{CLABTOOLKIT} enables researchers to construct more robust and readable analysis pipelines, aligning with the Findable, Accessible, Interoperable, and Reusable (FAIR) principles \cite{wilkinson_fair_2016} by ensuring that both data and workflows remain highly interoperable and reusable across different platforms. Furthermore, the package emphasizes interoperability with existing standards and libraries, allowing users to integrate it seamlessly into established workflows while reducing redundancy and improving reproducibility.\par
    
In this paper, we describe the design principles underlying \textsc{CLABTOOLKIT}, detail its core functionality, and demonstrate its utility through representative use cases. We aim to show that providing a well-structured layer of shared infrastructure can substantially reduce development overhead, improve code reliability, and ultimately accelerate scientific discovery in neuroimaging.\par

\section*{Software design and architecture}
\label{sec:sda}

\textsc{CLABTOOLKIT} comprises 19 Python modules grouped into the six categories summarized in Table~\ref{tab:clabtoolkit-overview}, totaling 13 classes, 234 methods, and 207 standalone functions (441 callable units). The five core data structures account for the largest share of class methods—\texttt{Parcellation} (41), \texttt{Surface} (35), \texttt{AnnotParcellation} (34), \texttt{Connectome} (25), and \texttt{Tractogram} (24)—reflecting the central role played by neuroanatomical primitives in the package, while the \texttt{BrainPlotter} class concentrates the 22 methods dedicated to multi-view rendering.\par

\begin{table}[htbp]
  \centering
  \caption{Overview of the \textsc{CLABTOOLKIT} package. The 19 Python modules are grouped by functional category. For each module, we report its key class(es)—where applicable—the total number of methods inside those classes, the number of standalone functions, and a short statement of purpose. }
  \label{tab:clabtoolkit-overview}
  \renewcommand{\arraystretch}{1.15}
  \begin{adjustbox}{max width=\textwidth}
  \begin{small}
  \begin{tabular}{@{}p{2.2cm} p{3.1cm} p{3.8cm} c c p{4.9cm}@{}}
  \toprule
  \textbf{Category} & \textbf{Module} & \textbf{Key class(es)} &
  \textbf{Methods} & \textbf{Functions} & \textbf{Purpose} \\
  \midrule
  \multirow{3}{*}{\makecell[c]{Data\\ Management}}
   & \texttt{bidstools}   & & &  23 & Parsing, validation and manipulation of BIDS datasets. \\
   & \texttt{imagetools}        & \texttt{MorphologicalOperations}  & 13 & 26 & Volumetric image manipulation and morphological operations. \\
   & \texttt{dicomtools}  & & &  7  & DICOM inspection and organization. \\

  \midrule
  \multirow{3}{*}{\makecell[c]{Utility \\\&\\ Infrastructure}}
   & \texttt{misctools}     & \texttt{SmartFormatter} & 1 & 53 & General-purpose utilities: parsing, file system, I/O, execution, inspection. \\
   & \texttt{colorstools}        & \texttt{ColorTableLoader}                   & 12 & 23 & Color maps, lookup tables and color harmonization. \\
   & \texttt{plottools}          &                                             &    & 8  & 2D plotting helpers (carpet plots, matrices, summaries). \\

  \midrule
  \multirow{4}{*}{\makecell[c]{Core \\ Entities}}
   & \texttt{parcellationtools} & \texttt{Parcellation}, \texttt{RegionTimeSeries}       & 41 &   & Volumetric atlases; region-wise time series and connectome generation. \\
   & \texttt{surfacetools}      & \texttt{Surface}                                       & 35 & 2  & Triangular mesh geometry and surface-based analyses. \\
   & \texttt{tracttools} & \texttt{Tractogram}      & 24 & 3 & Streamline I/O, resampling, filtering, smoothing and clustering. \\
   & \texttt{connectivitytools} & \texttt{Connectome} & 25 &  & Functional and structural connectome representation and analysis. \\

  \midrule
  \multirow{5}{*}{\makecell[c]{Specialized\\ Analysis}}
   & \texttt{dwitools}   & \texttt{DiffusionScheme} & 7  & 3 & DWI acquisition schemes, gradient tables manipulation, and scalar maps computation. \\
   & \texttt{segmentationtools} &                                   &    & 2  & Tissue and structure segmentation helpers. \\
   & \texttt{morphometrytools}  &                                   &    & 15 & Parsing and computation of volumetric and surface morphometric features. \\
   & \texttt{freesurfertools}   & \texttt{AnnotParcellation}, \texttt{FreeSurferSubject} & 34 & 16 & FreeSurfer I/O, surface annotations and subject-level wrapping. \\
   & \texttt{networktools}      &                     &    & 5 & Graph-theoretic metrics on connectivity matrices. \\
   
  \midrule
  \multirow{3}{*}{\makecell[c]{Pipelines \\ \&\\ Quality Control}}
   & \texttt{pipelinetools} &                         &   & 12 & Pipeline status. \\
   & \texttt{qcqatools}         &                                   &    & 6  & Quality control. \\
   &        &                                   &    &   & \\
  \midrule
  \multirow{3}{*}{\makecell[c]{Visualization}}
   & \texttt{pointstools}       & \texttt{PointCloud}                                    & 20 & 2  & 3D point clouds for landmarks, ROI centroids and electrodes. \\
   & \texttt{visualizationtools} & \texttt{BrainPlotter}                       & 22 & 1  & High-level multi-view rendering of brain objects. \\

  \midrule
  \textbf{Total} & \textbf{19 modules} & \textbf{13 classes} & \textbf{234} & \textbf{207} & \textbf{441 callable units} \\
  \bottomrule
  \end{tabular}
  \end{small}
  \end{adjustbox}
\end{table}

\subsubsection*{Ecosystem}
\label{sec:ecosystem}

\begin{figure}[htbp]
  \centering
  \includegraphics[width=0.95\linewidth]{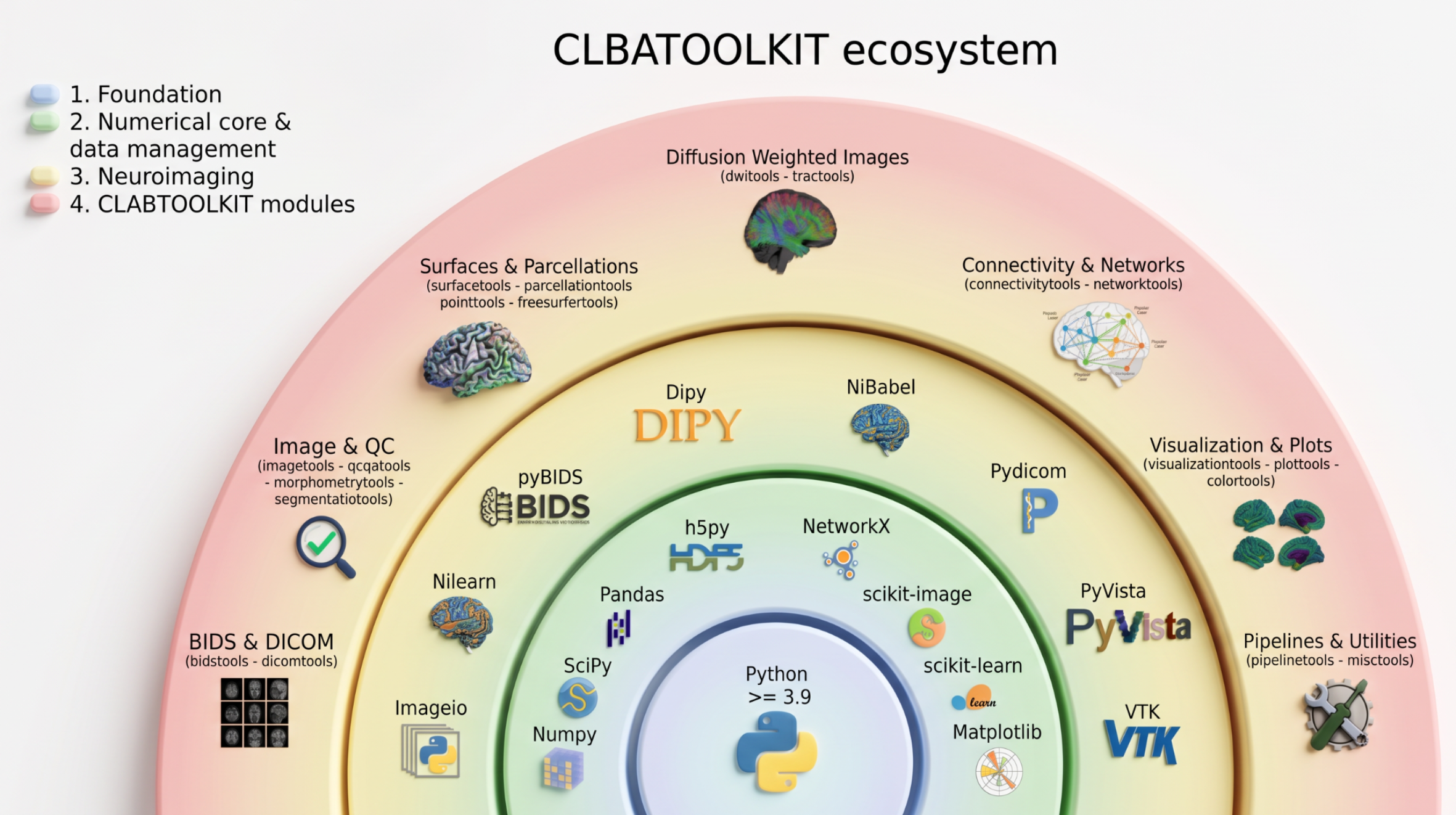}
  \caption{Overview of the \textsc{CLABTOOLKIT} ecosystem and its software dependencies.}
  \label{fig:ecosystem}
\end{figure}

\textsc{CLABTOOLKIT} is built on top of the scientific Python ecosystem and relies on established open-source libraries rather than re-implementing existing functionality (Figure~\ref{fig:ecosystem}). At its foundation is Python ($\geq 3.9$), which provides the runtime environment for the toolkit. The next layer contains the core libraries used for numerical computing, data handling, image processing, machine learning, graph analysis, and visualization, including NumPy \cite{harris_array_2020}, SciPy \cite{virtanen_scipy_2020}, pandas \cite{mckinney_data_2010}, h5py \cite{collete_hdf5_2008}, NetworkX \cite{hagberg_exploring_2008}, scikit-image \cite{van_der_walt_scikit-image_2014}, scikit-learn \cite{pedregosa_scikit-learn_2011}, and Matplotlib \cite{hunter_matplotlib_2007}. These libraries provide the basic data structures and computational tools used throughout the package.\par

A neuroimaging layer builds on these general-purpose libraries by providing support for medical image formats, BIDS datasets, diffusion MRI processing, and 3D visualization through NiBabel \cite{brett_nipynibabel_2023}, PyBIDS \cite{yarkoni_pybids_2019}, Nilearn \cite{contributors_nilearn_nodate}, DIPY \cite{garyfallidis_dipy_2014}, Pydicom \cite{mason_pydicom_2011}, and the VTK/PyVista\cite{sullivan_pyvista_2019} visualization framework. This layer allows \textsc{CLABTOOLKIT} to read and write common neuroimaging formats, interact with BIDS-compliant datasets, perform diffusion MRI analyses, and generate both two-dimensional and three-dimensional visualizations.\par

The outer layer corresponds to the \textsc{CLABTOOLKIT} modules. Rather than introducing new file formats or data structures, these modules combine functionality from the underlying libraries into a consistent set of tools for neuroimaging research. The modules are organized around common analysis tasks, including BIDS and Digital Imaging and Communications in Medicine (DICOM) management, image quality control, morphometric measurements, surface and parcellation analysis, diffusion MRI and tractography, connectivity and network analysis, visualization, and pipeline utilities. This organization allows users to move between different stages of an analysis using a common interface while maintaining compatibility with widely used neuroimaging software and community standards.\par

\subsubsection*{Architecture}
\label{sec:architecture}

\begin{figure}
  \centering
  \includegraphics[width=0.95\linewidth]{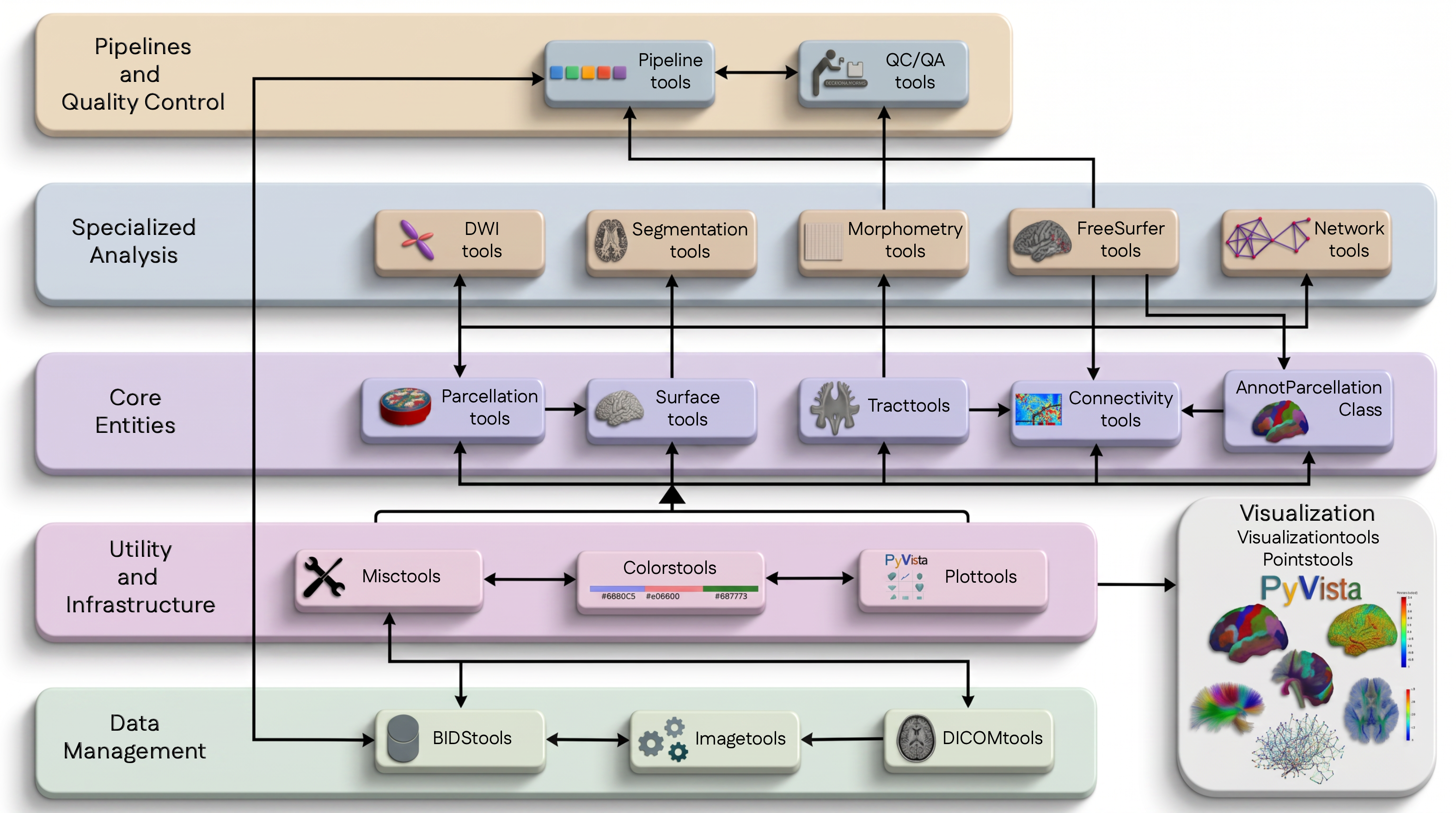}
  \caption{Simplified architecture diagram of the \textsc{CLABTOOLKIT} modules.}
  \label{fig:architecture}
\end{figure}

\textsc{CLABTOOLKIT} is functionally organized into six interconnected layers, each providing a distinct set of functionalities while collectively supporting a unified neuroimaging workflow (see Figure~\ref{fig:architecture}). \par

The data management layer supports data organization, renaming, format conversion, and interoperability across commonly used file types. The \texttt{dicomtools} module exports and organizes DICOM files into a subject/session/series hierarchy ready for BIDS conversion, with optional tar archival and header-level metadata access. The \texttt{bidstools} module enforces compliance with the Brain Imaging Data Structure by providing a formalized interface for entity parsing and manipulation, enabling consistent propagation of metadata throughout processing and into derived outputs. Foundational image operations—including binary morphology, affine-consistent coordinate transformations, and assembly of higher-dimensional volumes—are encapsulated in \texttt{imagetools}, which underpins the functionality of higher-level modules and ensures consistency in spatial image handling.\par

The utility and infrastructure layer underpins the rest of the toolkit. It groups three modules—\texttt{misctools}, \texttt{colorstools}, and \texttt{plottools}—that handle recurring low-level tasks: index parsing, filesystem operations, and progress tracking; the construction and manipulation of color maps for categorical and continuous data; and shared plotting primitives used by higher-level visualizations. Centralizing these operations removes ad hoc re-implementations from downstream modules and keeps figure aesthetics, parameter conventions, and I/O behavior consistent across pipelines.\par

The core entities layer contains the modules that define the core data structures on which the rest of the toolkit operates. It is built around five main classes—\texttt{Parcellation}, \texttt{Surface}, \texttt{AnnotParcellation}, \texttt{Tractogram}, and \texttt{Connectome}—each encapsulating a distinct neuroimaging entity together with the metadata and input/output routines required to manipulate it consistently (see the Core entities section).\par

To guarantee spatial consistency across the entire stack, every class enforces a common affine transformation convention, and the utility functions \texttt{vox2mm()} and \texttt{mm2vox()} provide bidirectional mapping between voxel indices and world-space millimeter coordinates. This unified treatment of geometry is essential for operations that jointly involve volumetric parcellations, surface annotations, point sets, streamlines, or connectivity matrices, and it is what allows the higher-level modules of \textsc{CLABTOOLKIT} to combine these primitives without ad hoc coordinate-system bookkeeping.\par

Extending these core capabilities, the specialized analysis layer comprises five modules that interface directly with established neuroimaging pipelines. The \texttt{dwitools} module provides utilities for working with diffusion-weighted imaging (DWI) data, such as removing volumes by index or b-value, extracting no-diffusion-weighted images, computing scalar maps from diffusion tensor eigenvalues, or representing and manipulating acquisition schemes (gradients and b-value tables) of a DWI dataset. The \texttt{segmentationtools} module performs atlas-based segmentation with user-supplied atlases to produce subject-specific tissue and regional labelings. The \texttt{morphometrytools} bridges the geometric representations and statistical workflows by computing a comprehensive set of regional descriptors, including surface-based scalar summaries, mesh-derived metrics (e.g., area, vertex count, Euler characteristic), volumetric statistics, parsed FreeSurfer outputs, and graph-theoretical connectivity measures. These features are provided as standard tables designed to plug directly into statistical analysis tools. The \texttt{freesurfertools} module formalizes access to FreeSurfer outputs through the \texttt{FreeSurferSubject} class, enabling structured querying of subject-level data and scalable generation of morphometric summaries across cohorts. The \texttt{parcellationtools} module is built on the \texttt{Parcellation} class, which contains methods to manipulate volumetric parcellations. Finally, the \texttt{networktools} module builds sparse CSR graphs from adjacency matrices, meshes, or edge lists, and supports graph metrics and connected-component extraction.\par

The pipelines and quality control layer includes modules for monitoring processing status and summarizing outputs from established pipelines such as FreeSurfer, as well as tools that leverage BIDS-compliant data organization to generate interactive web-based interfaces. These interfaces facilitate rapid quality control, visualization, and curation of neuroimaging datasets, enabling efficient inspection of large cohorts.\par

Finally, the visualization layer provides point clouds manipulation throughout the \texttt{pointstools} module as well as GPU-accelerated rendering capabilities through PyVista, enabling the generation of high-quality, multi-panel neuroimaging figures.\par

\subsection*{Configuration management}
\label{sec:configuration}

The framework employs a modular configuration architecture based on five external JSON files, enabling parameter customization without code modification. Three files govern data interpretation: \textit{bids.json} defines BIDS entity label mappings (e.g., ``sub'' $\rightarrow$ ``Participant''), standardizing file naming conventions used across the \texttt{morphometrytools} and \texttt{parcellationtools} modules; \textit{config.json} specifies physical units for a comprehensive set of morphometric and diffusion-derived metrics—ranging from cortical thickness (${mm}$) or surface area ($\mathrm{cm}^{2}$) to diffusion tensor indices such as mean diffusivity ($\text{mm}^2/\text{s}$)—ensuring consistent unit annotation throughout output tables; and \textit{stats\_mapping.json} provides structured mappings between toolkit-internal metric keys and the corresponding column names, alternate identifiers, segmentation codes, and unit divisors used in FreeSurfer statistics files (e.g., \texttt{aseg.stats} or \texttt{wmparc.stats}), to enable robust, format-agnostic parsing.\par

The remaining two files control anatomical grouping and visualization: \textit{lobes.json} defines region-to-lobe assignments for the Desikan–Killiany \cite{desikan_automated_2006}, including per-lobe color codes, supporting lobe-level aggregation in morphometric summaries; and \textit{viz\_views.json} specifies rendering parameters including camera positions, predefined viewing layouts, and lighting and reflection properties. All five files are loaded at runtime and can be extended or replaced by user-provided equivalents, allowing the toolkit to be adapted to custom atlases or alternative visualization conventions without modifying the source code.\par

\subsection*{Core Entities}
\label{sec:entities}

\subsubsection*{Parcellation}
\begin{figure}
    \centering
    \includegraphics[width=0.78\linewidth]{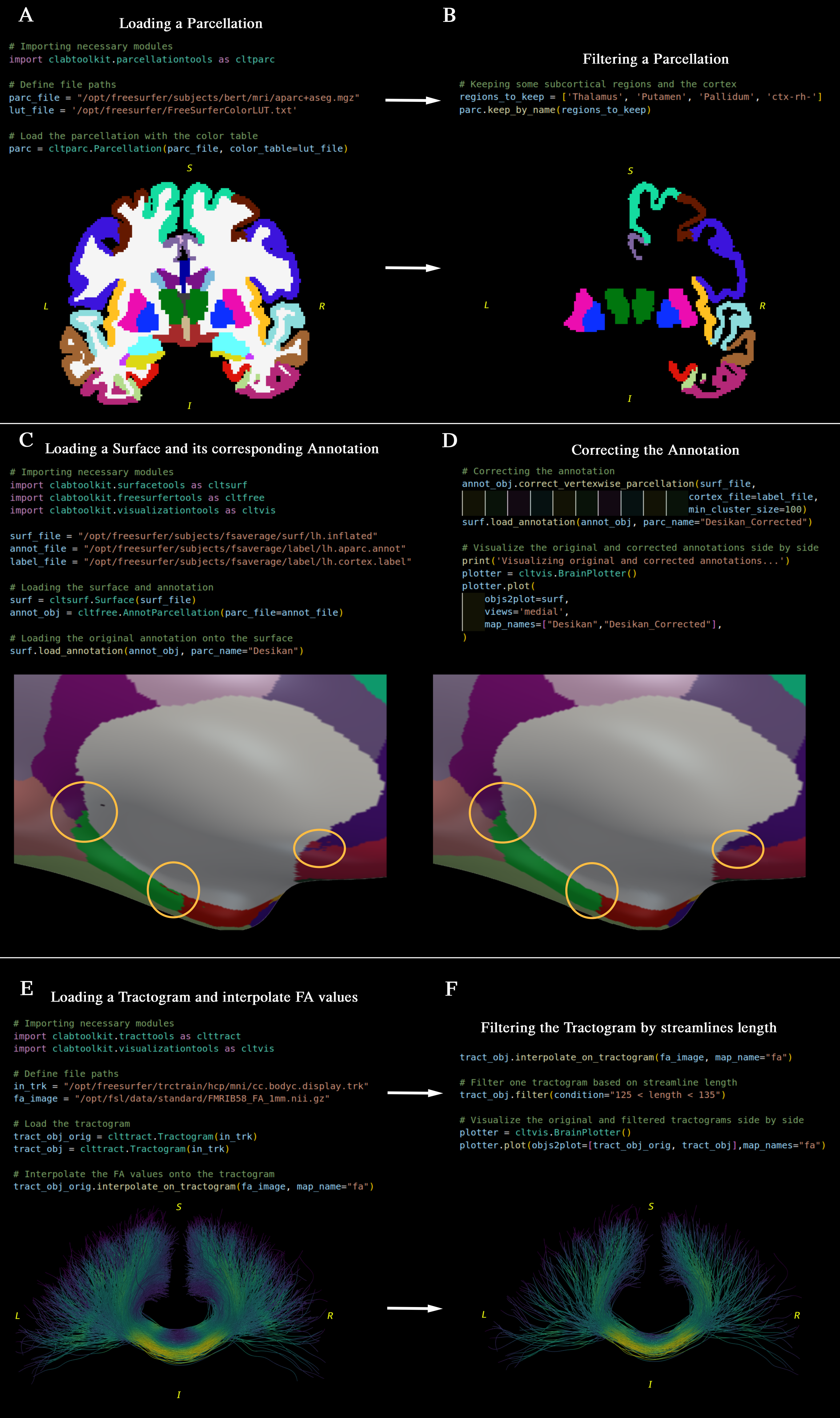}
      \caption{Overview of three core objects and operations in the workflow.
(A, C, E) illustrate the loading of the three objects: a \textit{Parcellation}, a \textit{Surface} with an \textit{AnnotParcellation}, and a \textit{Tractogram}, respectively. (B, D, F) Demonstrate representative methods applicable to each object.}
    \label{fig:objects}
\end{figure}

The \texttt{Parcellation} object represents a volumetric brain parcellation stored in NIfTI format (Figure~\ref{fig:objects}A). It can be instantiated from a NIfTI file or a NumPy array and optionally associated with a color lookup table provided as a tab-separated values (TSV) file, a lookup table (LUT) file, or a Python dictionary containing region identifiers, names, and RGBA color values. When a parcellation is loaded from disk, the corresponding lookup table is automatically loaded if available. The object stores the label volume together with its affine transformation, voxel resolution, and region metadata, including region names and color tables.\par

The class contains a set of operations for systematic manipulation of parcellations, including region selection by label or name (Figure~\ref{fig:objects}B), filtering and relabeling, aggregation of regions into coarser representations based on label ranges or naming conventions, and index remapping. It further supports deriving region-level properties from the underlying voxel representation, including volumetric estimates, adjacency relationships, and centroid coordinates. The \texttt{compute\_morphometry\_table} method provides a standardized interface for extracting per-region quantitative descriptors, returning a pandas DataFrame that includes volumetric measures and summary statistics (mean, median, minimum, maximum) of associated scalar maps, with optional filtering and export for downstream statistical analysis.\par

Input/output operations are designed to preserve interoperability with existing neuroimaging standards. The object supports writing parcellations to NIfTI format using BIDS-compliant naming conventions, as well as exporting to lookup tables and tab-separated values files. By unifying data representation, manipulation, and summarization within a single interface, the \texttt{Parcellation} object reduces the need for ad hoc implementations and facilitates reproducible, parcellation-based neuroimaging workflows.\par

Additionally, the \texttt{Parcellation} class serves as an entry point for region-wise analyses, providing direct methods to derive both time series and connectomes from any volumetric atlas. The \texttt{get\_regionwise\_timeseries()} method extracts an $N_{\mathrm{ROI}} \times T$ matrix—where $N_{\mathrm{ROI}}$ is the number of regions of interest (ROIs) and $T$ the number of time points—from a 4-D functional MRI (fMRI) volume (or a pre-loaded array) by aggregating voxel signals within each label using configurable summary statistics (mean, median, \textit{etc.}), with optional removal of dummy volumes and selection of regions by code or name. Building on this, \texttt{compute\_fc\_matrix()} generates a functional connectome by computing pairwise associations between regional time courses through a range of estimators (Pearson and Spearman correlation, partial correlation, mutual information) with built-in support for Fisher $z$-transformation, thresholding, and row normalization. For structural connectomes, the parcellation can be prepared for tractography through the \texttt{prepare\_for\_tracking()} method, and the resulting streamline counts or weights between every pair of regions can be assembled into a structural connectivity matrix. In all cases, the output is returned as a \texttt{Connectome} instance that automatically inherits the spatial coordinates, anatomical labels and color assignments of the originating parcellation, ensuring that functional and structural connectomes derived from the same atlas remain fully interoperable for downstream network analysis and visualization.

\subsubsection*{Surface}
Serving as a primary data structure within the toolkit, the \texttt{Surface} object establishes the framework for representing triangulated cortical surface meshes across all surface-based analytical workflows. It can be instantiated from FreeSurfer geometry files (e.g.\ \texttt{.pial}, \texttt{.white}, \texttt{.inflated}, \texttt{.sphere}), from explicit vertex coordinate and face connectivity arrays, or from existing PyVista mesh objects, allowing seamless integration with data originating from different processing pipelines. Internally, the mesh is stored as a PyVista \texttt{PolyData} object, which exposes vertices, faces, edges, and supports computing vertex normals and other geometric descriptors needed for downstream morphometric or topological analyses.\par

Beyond geometry, the \texttt{Surface} object acts as a container for any quantity defined on the cortical surface. Vertex-wise or region-wise scalar maps can be loaded from FreeSurfer overlays (\texttt{.mgh}, \texttt{.thickness}, etc.), NumPy arrays, CSV files, or pandas DataFrames, and the input format is automatically detected. When the data are region-wise, an \texttt{AnnotParcellation} object (Figure~\ref{fig:objects}C) is used to propagate regional values onto individual vertices. Multiple maps can coexist on the same surface as named overlays, which can be listed, queried, activated, or removed independently, enabling the exploration of several modalities on a common geometric substrate. Cortical parcellations such as FreeSurfer annotation (\texttt{.annot}) files can be similarly attached to the mesh and queried at the regional or vertex level. Volumetric scalar maps can also be projected onto the surface, bridging volume- and surface-based representations within a single object.\par

The \texttt{Surface} object provides operations commonly used in surface-based analyses, including the extraction of disconnected components, merging of multiple surfaces, and selection of vertices belonging to specific parcellation regions. Surface geometry, overlays, and annotations can be exported in several formats, including FreeSurfer, Wavefront OBJ, and PyVista. The object is also integrated with the visualization framework, supporting multi-view displays, custom colormaps, and the simultaneous visualization of surface geometry, overlays, and parcellations for data inspection and quality control.\par

\subsubsection*{AnnotParcellation}
The \texttt{AnnotParcellation} object (Figure~\ref{fig:objects}C) is the primary data structure for representing vertex-wise cortical parcellations, and it is involved in all region-based analyses within the toolkit. Each parcellation is internally encoded as a vertex-wise array of integer labels paired with a region color table and a set of region names, such that every cortical vertex is uniquely mapped to an anatomical region, an RGBA color, and a label. The object can be constructed from FreeSurfer annotations (\texttt{.annot}) or GIFTI label files (\texttt{.gii}), all of which are normalized to a common internal representation upon loading. In addition, the object can be initialized directly from in-memory arrays, allowing parcellations produced by clustering algorithms or machine-learning models to be handled identically to those read from disk.\par

The \texttt{AnnotParcellation} object provides functions commonly used in surface-based neuroimaging analyses. Regions can be queried by name or label to obtain the corresponding color, identifier, and vertex count. A summary with these statistics can be generated for the entire parcellation. The object includes methods to reassign isolated or disconnected vertex clusters based on surface neighbourhood information (Figure~\ref{fig:objects}D) and to interpolate unlabeled vertices within the cortical surface. For multi-scale analyses, regions can be grouped into larger anatomical units using predefined schemes, such as the Desikan–Killiany lobar hierarchy, or custom mappings provided as JSON files.\par

The \texttt{AnnotParcellation} object can be exported to the FreeSurfer \texttt{.annot} or \texttt{.gii} format and the metadata can be saved as a TSV file containing annotation identifiers, parcellation indices, region names, and hexadecimal color codes. The object is designed to work with the \texttt{Surface} object, which provides the surface geometry required to map region-level data to vertices, retrieve vertices associated with a given region, and visualize cortical parcellations and derived measurements.\par

\subsubsection*{Tractogram}
The \texttt{Tractogram} object (Figure~\ref{fig:objects}E) provides a representation of white matter tractography data and serves as the central data structure for streamline-based analyses in the toolkit. It supports the two most widely used file formats, TrackVis (\texttt{.trk}) and MRtrix (\texttt{.tck}), and can also be instantiated directly from a nibabel \texttt{Tractogram} object, ensuring interoperability with the broader Python neuroimaging ecosystem. Internally, streamlines are stored as a nibabel \texttt{ArraySequence}, along with the affine matrix and file header to preserve the spatial reference frame of the source dataset.\par

This object offers a set of operations required to prepare streamlines for analysis. Streamlines can be resampled to a fixed number of points to homogenize the sampling density across bundles, smoothed via Gaussian filtering to reduce reconstruction noise, filtered according to user-defined criteria (Figure~\ref{fig:objects}F), and randomly subsampled to obtain lighter representations for visualization. Descriptive statistics, including streamline length distributions and global summaries of the tractogram, can be computed on demand, and multiple tractograms can be merged into a single object while preserving their associated metadata.\par

A key feature of the \texttt{Tractogram} object is its ability to act as a container for any quantity defined along or per streamline. Scalar values from NIfTI images (e.g.\ fractional anisotropy, mean diffusivity, or any other voxel-wise map) can be projected onto the tractogram through an interpolation at each streamline point. The resulting values can be retained at full resolution as \texttt{data\_per\_point} or aggregated per streamline using statistical reductions such as mean, median, minimum, or maximum and stored as \texttt{data\_per\_streamline}. Multiple named maps can coexist on the same tractogram and can be listed, queried, and combined to support future multimodal analyses.\par

The object also integrates streamline clustering through the QuickBundles \cite{garyfallidis_quickbundles_2012} algorithm provided by DIPY, enabling the identification of coherent bundles and supporting cluster-level operations such as centroid extraction, bundle labelling, and selective retrieval of streamlines from individual clusters. Tractograms can be exported back to both \texttt{.trk} and \texttt{.tck} formats, optionally with compression, and connect to the toolkit's visualization engine to render streamlines together with their associated scalar maps, cluster assignments, and anatomical context.\par

\subsubsection*{Connectome}
The \texttt{Connectome} object represents brain connectivity data and constitutes the core data structure for network-based analyses within the toolkit. A connectome is stored as a square $N_{\mathrm{ROI}} \times N_{\mathrm{ROI}}$  NumPy array, where $N_{\mathrm{ROI}}$ is the number of brain regions defined by a given parcellation, together with the metadata required to interpret the network in an anatomically meaningful way. Each region is described by an integer index, a human-readable name, an RGBA color suitable for consistent visualization across figures, and optionally a three-dimensional anatomical coordinate that locates the region in a reference space. The object also records the type of connectivity it encodes (e.g.\ \texttt{structural}, \texttt{functional}), enforces basic consistency checks such as matrix symmetry, and keeps track of the affine transformation linking regional coordinates to the underlying image space, so that the connectome can be overlaid on volumetric or surface-based representations without ambiguity.\par

The \texttt{Connectome} object provides methods for standard connectivity analyses. Global network measures such as density, mean connectivity, dispersion, and extreme values can be computed directly from the object. Connectivity matrices can be thresholded using user-defined criteria to extract salient edges. Subnetworks can be generated by clustering using community detection methods or by selecting regions of interest, returning a new \texttt{Connectome} instance while preserving associated metadata. Region attributes, including names, colors, and coordinates, are accessed through dedicated methods, and missing values are assigned default entries to maintain a consistent internal state.\par

For storage and interoperability, connectomes and their metadata, coordinates, and affine transformation are saved to and loaded from HDF (\texttt{.h5}) files, optionally with compression. The object also provides basic visualization tools, including matrix plots, circular layouts, and three-dimensional node-and-edge representations using PyVista, where node size reflects regional connectivity and edges encode connection strength. Visualizations can be exported to standard image formats for inspection and figure generation.\par

\subsection*{Core Modules}
\label{sec:modules}

\subsubsection*{\texttt{parcellationtools}}
The \texttt{parcellationtools} module provides a framework for managing and analyzing volumetric brain parcellation data, and it organizes this functionality around the central \texttt{Parcellation} class, which serves as the container for volumetric atlases throughout the toolkit.\par

The module contains a set of label-space operations that support both exploratory and systematic analyses. Region filtering is available through both label codes and name sub-string matching, allowing users to keep or remove specific structures in a single call. Regions can also be grouped into coarser units by code or by name to obtain custom parcellations at multiple levels of granularity. Additionally, label codes can be rearranged, harmonized across parcellations, or replaced according to user-defined mappings, ensuring consistent indexing when combining data from different atlases. Region adjacency can be computed directly from the volumetric labels to obtain neighbourhood graphs suitable for graph-theoretical analyses. For spatial operations, the module supports mask application to restrict a parcellation to an anatomical region of interest, as well as the masking of arbitrary images by the parcellation itself. Region centroids can be computed in both voxel and millimeter coordinates with optional Gaussian smoothing and morphological closing, and the results are exported as structured DataFrames or TSV files.\par

For functional and morphometric applications, the module extracts region-wise time series from 4D NIfTI data, supporting both mean and median aggregation via either NumPy or Nilearn backends, and returning the result as a dedicated \texttt{RegionTimeSeries} object that bundles the extracted signals with their associated regional metadata. From this representation, functional connectivity matrices can be computed directly, providing a seamless transition from voxel-level data to the \texttt{Connectome} representation. Morphometric quantification is supported through two dedicated methods: \texttt{compute\_volume\_table}, which calculates the volume of each parcellation region in cubic centimeters, and \texttt{compute\_morphometry\_table}, which additionally overlays arbitrary scalar maps (e.g., microstructure or quantitative MRI maps) onto the parcellation to derive per-region summary statistics.\par

Surface mesh generation is also possible through a marching cubes \cite{lorensen_marching_1987} pipeline with Taubin \cite{taubin_curve_1995} smoothing, hole filling, and optional merging of all region meshes into a single annotated surface, exportable in FreeSurfer, VTK, PLY, STL, and OBJ formats, supporting both visualization and downstream surface-based processing.\par

Finally, parcellations can be combined via direct overlay or label-offset appending, allowing multiple atlases or hand-curated regions to be merged into a single, consistent object without manual relabelling. The full object state can be serialized to a NIfTI image together with its lookup table as a sidecar file, exported to HDF5, or written as standalone color tables in standard formats, so that parcellations produced or modified within the toolkit remain fully interoperable with external neuroimaging software and analysis frameworks.\par

\subsubsection*{\texttt{misctools}}
The \texttt{misctools} module provides the general-purpose utility layer on which the rest of the toolkit relies. Rather than exposing a single coherent abstraction, it gathers a collection of small, self-contained functions that recur across neuroimaging analyses—parsing user input, manipulating filenames, handling tabular metadata, introspecting code, or wrapping external software—and offers them through a uniform API. By centralizing these helpers in a single module, we both eliminate code duplication across the toolkit's other components and ensure consistent behavior (error handling, type coercion, BIDS-aware parsing) regardless of the calling context. For clarity, the utilities can be grouped into six thematic blocks: index and range handling, list and string operations, file system operations, data input/output, execution helpers, and data inspection.\par

\textit{Index and range handling} addresses the recurrent need to translate human-friendly specifications—such as those provided on the command line or in configuration files—into machine-usable arrays. The function \texttt{build\_indices} accepts heterogeneous inputs (integers, tuples, NumPy arrays, and strings using dash, colon, or comma notation, e.g.\ \texttt{"1-5,8,10:12"}) and returns a single sorted, deduplicated list of integers. For data-driven selection, \texttt{get\_indices\_by\_condition} and \texttt{get\_values\_by\_condition} evaluate arbitrary logical expressions (e.g.\ \texttt{"bvals > 1000"}) against named arrays and return, respectively, the matching positions or the matching values; this allows users to filter diffusion shells, time points, or atlas regions without leaving the high-level API.\par

\textit{List and string operations} target the diverse nomenclature typically encountered in neuroimaging metadata: sub-string queries with OR/AND semantics and optional case sensitivity, regex-based or literal cleanup, collapse of repeated characters generated by automated label exports, and a label-correction pipeline that chains stripping, regex replacement, case conversion and affix insertion in a deterministic order. These primitives make it straightforward to harmonize region labels, subject identifiers, or filenames in a reproducible way.\par

\textit{File system operations} encapsulate the recursive directory traversal, batch renaming, and pruning patterns that are otherwise repeatedly re-implemented across neuroimaging pipelines. The module can enumerate files recursively under arbitrary OR/AND filename filters, apply batch string substitutions across entire directory trees, and prune empty directories bottom-up. Destructive operations expose a \emph{simulation} mode that previews the resulting changes without touching the disk, which is particularly valuable when curating large datasets or reorganising legacy archives.\par

\textit{Data input/output} utilities support the simple file formats that appear throughout neuroimaging workflows. The module reads and writes JSON configuration files, loads delimiter-separated tables, automatically detects the separator, keeps the original BIDS entity types, and merges nested dictionaries using rules that the user can set for conflicting keys and list values. These tools let the higher-level components load metadata, sidecar files, and parameter sets without having to re-implement format detection or merge logic.\par

\textit{Execution helpers} enable the toolkit to run command-line programs across different environments with a unified interface. Given a command, the helper automatically wraps it for execution on the local machine, inside a Docker container, or within a Singularity container. It intelligently determines which folders to share with the container by analyzing file paths in the command. By abstracting away the underlying container system, this single helper allows higher-level pipeline modules to remain agnostic to whether Docker, Singularity, or no container system is installed on the host.\par

Finally, a set of \textit{data inspection} tools facilitates interactive use and documentation. These tools can list all classes, functions, and attributes of any toolkit component, complete with full signatures, and automatically adapt their output format for a terminal or a notebook. They can also display the contents of HDF5 files as a color-coded tree, showing each dataset's shape, data type, size, and attributes, as well as search a given module's functions by keyword with ranked matches.\par

\subsubsection*{\texttt{visualizationtools}}
The \texttt{visualizationtools} module provides a high-level visualization interface for neuroimaging objects, built on top of PyVista and exposed through the \texttt{BrainPlotter} class. A \texttt{BrainPlotter} is initialized from a JSON configuration file that defines available camera views, multi-view layout templates, figure styling (background color, fonts, mesh shading parameters), and per-object rendering options for surfaces, tractograms, and point clouds. A default configuration is bundled with the toolkit. It can be extended or replaced either at construction time or during plotting, allowing the same scene to be re-styled without rewriting the calling code.\par

Three entry points cover the most common use cases. The first accepts any combination of \texttt{Surface}, \texttt{Tractogram}, and \texttt{PointCloud} objects, arranging them across one or more named views—such as lateral, medial, dorsal, ventral, rostral, caudal, or predefined multi-view layouts—in a configurable grid. It supports multiple data maps and objects simultaneously, featuring per-map color map selection, value clamping, masking of out-of-range values, and individual or shared color bars positioned to the right or below the figure. The second is a dedicated bilateral interface where the left and right objects are passed separately; here, each panel is automatically labeled by hemisphere, and the color map range is computed jointly across both hemispheres. The third is a free-form compositor that accepts a heterogeneous list of objects alongside a per-object configuration (map name, color map, value limits, opacity) to produce layered scenes where surfaces, tractograms, and point clouds can be combined. Tractograms can be rendered as colored lines or as tubular meshes with configurable radii and side counts, while point clouds can be drawn as flat points or spheres.\par

All three entry points can render scenes in an interactive terminal window (optionally non-blocking via a background thread), embed them directly inside a notebook, or save them to an image or HTML file without opening a display. Camera positions across subplots sharing the same view are linked; rotating one panel updates all matching panels simultaneously, making side-by-side comparisons across subjects or hemispheres straightforward.\par

Finally, figure styling can be modified at runtime, switched to a named theme, previewed before application, and written back to the configuration file. This allows the visual identity of a study to be defined once and reused across all figures produced with the toolkit.\par

\section*{Summary of Contributions}
\label{sec:contributions}

\textsc{CLABTOOLKIT} is not a neuroimaging processing pipeline, but rather a utility framework designed to simplify and standardize the ancillary tasks that surround neuroimaging data analysis. One of its central contributions is a comprehensive color management system, which provides tools for creating, manipulating, and converting color tables, enabling consistent and flexible color assignment across parcellations and visualizations. Closely related is the toolkit's parcellation handling functionality, which allows researchers to load, filter, group, and relabel brain parcellations, compute regional volumes, and manage associated lookup tables in a unified interface. The toolkit also facilitates the construction of structured indexes and tabular outputs from neuroimaging derivatives, streamlining the preparation of region-of-interest data for statistical analyses in standard environments such as R or Python statistical libraries. On the FreeSurfer side, the toolkit extends native functionality by manipulating cortical annotation files, correcting parcellation artifacts, and converting between surface file formats. A further contribution is the integrated surface visualization engine, which enables researchers to render multiple brain surfaces simultaneously with multiple overlaid scalar maps—such as cortical thickness, annotations, or connectivity metrics—using configurable multi-view layouts, without requiring expertise in dedicated rendering software. The toolkit also provides a rich set of tractography utilities that go beyond simple streamline loading: researchers can interpolate scalar maps along streamlines, assign and store per-streamline or per-point attributes directly onto tractograms, resample streamlines to a fixed number of points, and perform bundle-level operations such as clustering and statistical summarization. These capabilities are particularly valuable for white matter microstructure studies, where mapping diffusion metrics onto specific tracts and extracting along-tract profiles are common but technically demanding tasks. Together, these features reduce the time spent on repetitive, error-prone bookkeeping tasks, allowing researchers to move more efficiently from processed neuroimaging data to scientific interpretation and reporting.\par

\begin{figure}
  \centering
  \includegraphics[width=1\linewidth]{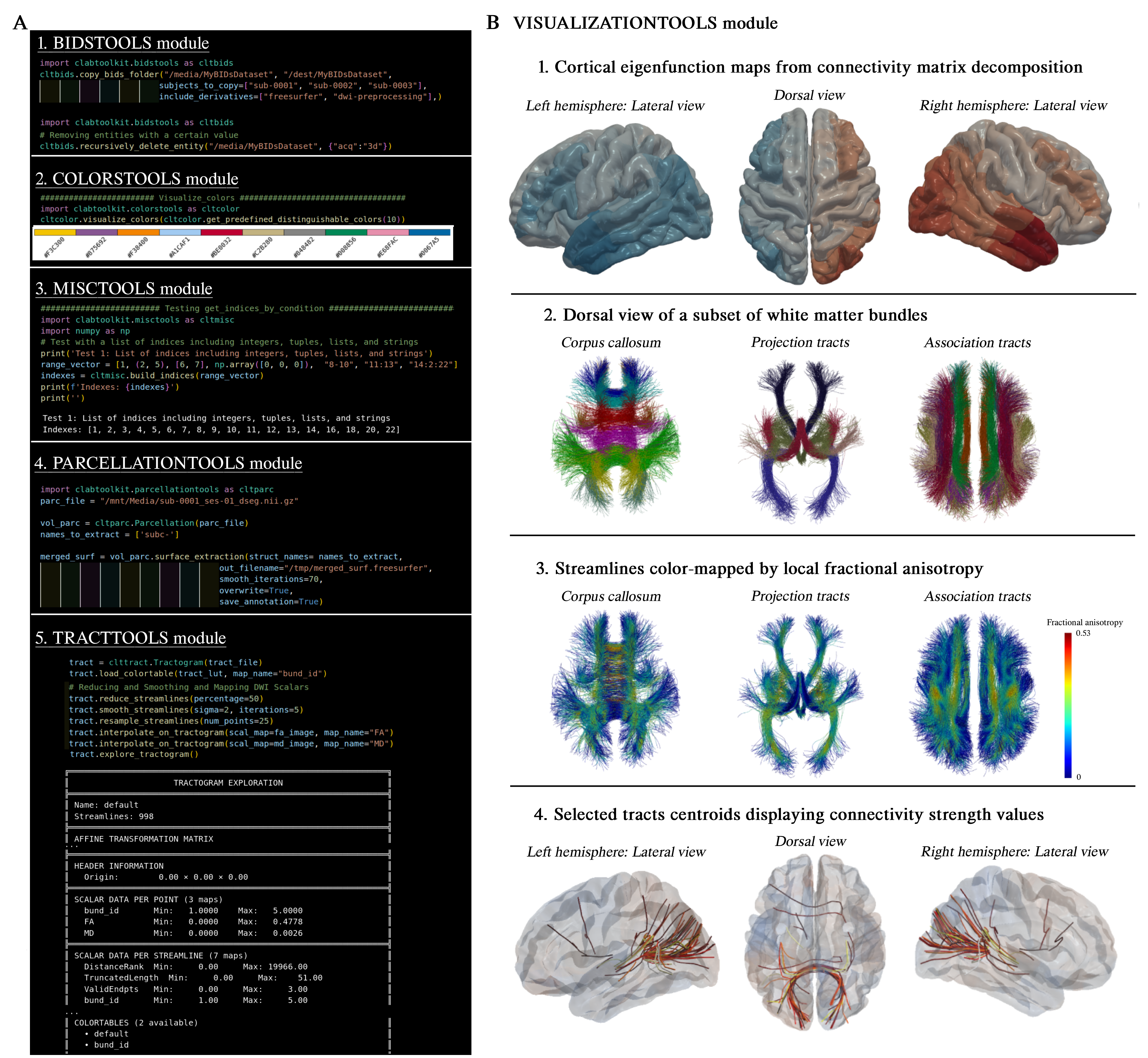}
  \caption{Code examples and visualization capabilities of \textsc{CLABTOOLKIT}. (A) Code examples demonstrating the usage of five core modules within \textsc{CLABTOOLKIT}. (B) Visualization outputs from the \texttt{visualizationtools} module demonstrating its flexibility to display vertex-wise, point-wise, and region-wise maps as well as tractograms. \textbf{Note:} New visualization scenes can be created by combining these examples with different data modalities and rendering parameters.}
  \label{fig:visualization}
\end{figure}

Figure~\ref{fig:visualization} illustrates representative use cases for \textsc{CLABTOOLKIT}, demonstrating how concise code can yield immediately actionable analysis. Specifically, Figure~\ref{fig:visualization}A highlights practical applications across five distinct modules. These include stripping a target entity with a certain value from the file names of the files contained by the BIDS dataset (Figure~\ref{fig:visualization}A.1), as well as generating and displaying ten easily contrastable colors (Figure~\ref{fig:visualization}A.2). Additionally, the toolkit can construct indices from diverse lists containing integers, tuples, nested lists, NumPy arrays, or comma-separated expressions (Figure~\ref{fig:visualization}A.3). For parcellation analysis, it leverages the marching cubes algorithm to extract boundaries for any parcellation regions matching a specific text string (Figure~\ref{fig:visualization}A.4). Finally, Figure~\ref{fig:visualization}A.5 illustrates a variety of direct manipulations performed on a \texttt{Tractogram} object.\par

Figure~\ref{fig:visualization}B shows the rendering outputs produced by the \texttt{visualizationtools} module for different cases: vertex-wise scalar maps overlaid over a \texttt{Surface} (Figure~\ref{fig:visualization}B.1), bundles, and point-wise fractional anisotropy values overlaid on a \texttt{Tractogram} (Figure \ref{fig:visualization}B.1 and \ref{fig:visualization}B.2, respectively) and a scene containing a transparent \texttt{Surface} and some \texttt{Tractogram} centroids (\ref{fig:visualization}B.4).\par

\subsection*{Configurability without code modification}
\label{sec:configurability}

The JSON-based configuration system (see the Configuration section) allows users to adapt the toolkit to new atlases, scanner conventions, or visualization styles without editing the source code. In practice, replacing \textit{lobes.json} suffices to support a custom lobar grouping; substituting \textit{stats\_mapping.json} adapts the FreeSurfer parser to alternative or extended statistics files; and editing \textit{viz\_views.json} introduces new camera presets or multi-view layouts that propagate to every rendering call. Together with the runtime style overrides exposed by \texttt{BrainPlotter}, these mechanisms keep the visual identity of a study consistent across all figures while keeping the underlying analysis code unchanged.\par

\textsc{CLABTOOLKIT} relies on several established Python libraries for scientific computing and neuroimaging. Core numerical and array operations are handled by NumPy \cite{harris_array_2020} and SciPy \cite{virtanen_scipy_2020}, while tabular data management uses pandas \cite{mckinney_data_2010}. Neuroimaging-specific I/O is provided by NiBabel \cite{brett_nipynibabel_2023} for NIfTI and MGH file formats and Pydicom \cite{mason_pydicom_2011} for DICOM data. Higher-level neuroimaging operations leverage Nilearn \cite{contributors_nilearn_nodate}, and diffusion-weighted imaging (DWI) processing relies on DIPY \cite{garyfallidis_dipy_2014}. Image processing routines are built on scikit-image \cite{van_der_walt_scikit-image_2014}, and 3D surface visualization is enabled by PyVista \cite{sullivan_pyvista_2019}. A full list of dependencies is provided in \texttt{requirements.yaml} in the repository.

\subsection*{Comparison with Related Work}

Several Python tools exist for neuroimaging visualization and analysis, each addressing specific aspects of the workflows that \textsc{CLABTOOLKIT} aims to integrate. Nilearn\cite{contributors_nilearn_nodate} offers a broad set of utilities including atlas handling, parcellation, signal extraction, and surface plotting, making it the most widely used general-purpose neuroimaging library; however, its color and lookup table management relies on standard Matplotlib color maps, its surface visualization is largely supplementary to its volumetric focus, and it provides no tractography utilities. BrainSpace\cite{vos_de_wael_brainspace_2020} is well suited for gradient-based cortical analyses and multi-map surface visualization, but is limited to continuous data representations and lacks discrete parcellation manipulation, lookup table management, and any tract-based functionality. Surfplot provides a clean interface for generating publication-ready cortical surface figures with flexible multi-view layouts, but is restricted to static two-dimensional figures and does not support parcellation editing, custom lookup table creation, or tractogram manipulation. Pycortex\cite{gao_pycortex_2015} excels at interactive web-based visualization of fMRI data projected onto cortical surfaces, but is specialized for that use case and offers no parcellation or tractography tools. On the diffusion MRI side, scilpy\cite{renauld_tractography_2026} provides tract profile extraction and streamline-level scalar mapping, but neither addresses surface-based visualization, parcellation manipulation, nor color table management. FURY\cite{garyfallidis_fury_2021}, the visualization backend of DIPY, enables high-performance streamline rendering with scalar color mapping, but does not include parcellation or tractogram manipulation because it is mainly a tool dedicated to general scientific data visualization. In contrast, \textsc{CLABTOOLKIT} uniquely combines lookup table and color management, parcellation manipulation, multi-surface multi-map visualization, and tractography utilities—including streamline interpolation, scalar mapping onto tractograms, and streamline resampling—within a single, cohesive framework, filling a gap that currently requires combining multiple tools with different interfaces and conventions.

\subsection*{Limitations}

While \textsc{CLABTOOLKIT} provides a comprehensive framework for neuroimaging data processing and analysis, several limitations should be acknowledged. First, certain functionalities depend on external neuroimaging tools—specifically FreeSurfer and ANTs—which must be independently installed and properly configured in the user's environment, either natively or via container runtimes (Docker or Singularity). This introduces additional setup complexity and potential version-specific incompatibilities. Second, the toolkit is primarily optimized for structural MRI and diffusion MRI workflows; support for functional MRI is limited and largely delegated to Nilearn, with no built-in temporal dynamics analysis. Third, the statistical analysis capabilities are currently confined to descriptive regional statistics; inferential statistical modeling is not natively supported and must be performed with external libraries. Finally, while BIDS entity parsing and naming conventions are well supported, no formal BIDS dataset validation is performed, and users are responsible for ensuring their datasets conform to the BIDS specification before processing.

\section*{AI disclousure}
AI assistance was used during the development of \textsc{CLABTOOLKIT} and the preparation of this manuscript. Specifically, GitHub Copilot (autocomplete mode within VS Code) and the Claude Code were used to support source code development, primarily for minor features and portions of the test suite. Additional AI tools were used for language editing to improve clarity and wording in selected parts of the manuscript. All AI-generated or AI-modified code and text were thoroughly reviewed and proofread by the authors, and all AI-assisted code was extensively tested before integration.

\section*{Data and code availability}
\textsc{CLABTOOLKIT} is freely available under the Apache Software License 2.0 at \url{https://github.com/connectomicslab/clabtoolkit} and can be installed directly via the Python Package Index (PyPI) at \url{https://pypi.org/project/clabtoolkit/}.
Documentation is hosted at \url{https://clabtoolkit.readthedocs.io}. 

\vspace{0.8cm}

\section*{Acknowledgements}
\addcontentsline{toc}{section}{Acknowledgements}

\noindent\textbf{Funding:} This work was supported by the Swiss National Science Foundation (SNSF) Sinergia grant \href{https://www.mysnf.ch/grants/grant.aspx?id=0e084d3a-46ce-484f-aaf4-b6ebdbfe1d5b}{CRSII5\_209470}.

\vspace{0.8cm}

\section*{Conflict of interest}
\addcontentsline{toc}{section}{Conflict of interest}
The authors declare no conflict of interest.

\vspace{0.8cm}

\section*{Author Contributions}
\addcontentsline{toc}{section}{Author Contributions}

\noindent
\textbf{YA:} Conceptualization, Methodology, Software, Writing – original draft.\\
\textbf{NH:} Writing – review \& editing.\\
\textbf{PH:} Writing – review \& editing.\\

\newpage
\printbibliography[heading=bibintoc, title={References}]


\end{document}